\documentclass[german,english]{bvm2019}

\title{Augmented Mitotic Cell Count using Field Of Interest Proposal}

\author{Marc Aubreville$^1$, Christof A. Bertram$^2$, Robert Klopfleisch $^2$, Andreas Maier $^1$}

\authorrunning{Aubreville et al.}

\institute{%
$^1$Pattern Recognition Lab, Computer Sciences, Friedrich-Alexander-Universit\"at Erlangen-N\"urnberg\\
$^2$Institute of Veterinary Pathology, Freie Universit\"at Berlin, Germany\\}

\email{marc.aubreville@fau.de}

\begin{document}

\selectlanguage{english}

\maketitle

\begin{abstract}
Histopathological prognostication of neoplasia including most tumor grading systems are based upon a number of criteria. 
Probably the most important is the number of mitotic figures which are most commonly determined as the mitotic count (MC), i.e. number of mitotic figures within 10 consecutive high power fields. Often the area with the highest mitotic activity is to be selected for the MC. 
However, since mitotic activity is not known in advance, an arbitrary choice of this region is considered one important cause for high variability in the prognostication and grading.
\\In this work, we present an algorithmic approach that first calculates a mitotic cell map based upon a deep convolutional network. This map is in a second step used to construct a mitotic activity estimate. Lastly, we select the image segment representing the size of ten high power fields with the overall highest mitotic activity as a region proposal for an expert MC determination. We evaluate the approach using a dataset of 32 completely annotated whole slide images, where 22 were used for training of the network and 10 for test. We find a correlation of r=0.936 in mitotic count estimate. 
\end{abstract}

\section{Introduction}
One important aspect of tumor prognostication in human and veterinary pathology is the proliferative rate of the tumor cells, which is assumed to be correlated with the density of cells undergoing divison (mitotic figures) in a histology slide\cite{elston1991pathological} and is applied as a criterion in almost all current tumor grading systems. However, mitotic activity is known to have large inter-observer variances~\cite{PoulBoiesenParOlaBendahlLola:2009kv}, which consequentially strongly affects the histological grade assigned. One reason might be that the classification between mitotic and non-mitocic cells is not clearly defined and varies across labs, schools and even individuals~\cite{Meuten:2016jh,bertram2018validation}. Another important reason for this is, that the distribution of mitotic cells in the slide is usually sparse with local changes in density across the specimen. In clinical practice, this sampling problem is dealt with by counting mitotic figures in ten fields of view at a magnification of 400$\times$ (high power fields, HPF), resulting in the mitotic count~(MC). However, as shown previously \cite{Bonert:2017go}, especially for low to borderline mitotic counts, semi-random selection of those ten high power fields is not sufficient for a reproducible MC determination. While examining larger areas would improve on this, it is not the method of choice given limited time budgets in pathology labs. As of this writing, completely algorithmic approaches for mitotic activity estimation lack the sensitivity and specificity that would be required to achieve clinical applicability. Further, purely algorithmic outcomes may be subject to hesitation from the pathology side, since automatic solutions that are not easily comprehensible for the medical expert, such as deep learning networks, may not be robust.

In this paper, we present an algorithmic approach that proposes a region of the area of 10 high power fields that is assumed to have the highest mitotic count within the slide. This has two positive aspects: While we still rely on the expertise of a pathologist to assess the actual mitotic activity, we limit the focus area to a defined field of interest in the image. Further, as this algorithmic answer will always be equal for the same image, it will allow us to differentiate the true inter-observer variance in an optimal setting when the area on the slide is already fixed. This region proposal will serve as an augmentation to the pathology expert.

\section{Material}

For this work, we annotated 32 histology slides of canine cutaneous mast cell tumors dyed with standard hematoxylin and eosin stain. The slides were digitized using a linear scanner (Aperio ScanScope CS2, Leica Biosystems, Germany) at a magnification of $400 \times$ (resolution: 0.25\,$\frac{\mu m}{\textrm{px}} $). Contrary to popular other publicly available mitosis data sets, we did not pre-crop the whole slide images (WSI) but include all parts of the slide, including borders, which we consider important for a general applicability of the framework. All slides have been annotated by two pathologists using the open source annotation software SlideRunner \cite{univis91884757}. Out of all cells annotated as mitotic figure, we only use those where both observers agreed upon being a mitotic figure. 
We arbitrarily chose 10 slides to be the test set, and 22 to be used for the training process. The data set includes slides of low, medium and high mitotic activity in both training and test set. In total, 45,811 mitotic cells have been annotated on all slides. To the best of our knowledge, this data set is unprecedented in size for any mitotic cell task and may serve as basis for many algorithmic improvements to the field.

\section{Methods}

A significant number of algorithmic approaches for mitosis detection have emerged very recently, most based on deep convolutional networks \cite{Ciresan:2013up,akram2018leveraging,paeng2017unified}, making use of transfer learning \cite{Li:2018ce} and hard-negative example mining \cite{akram2018leveraging,Ciresan:2013up}. Typically, these algorithms use a two-stage approach, where in the first stage multiple regions of interest are detected and in a second stage classification is done according to being a mitotic figure or not. However, as also stated in the TUPAC challenge, automated identification of mitoses is only an intermediate step in tumor grading. F1-scores of up to 0.652 \cite{paeng2017unified} have been achieved on the TUPAC challenge test data set. Current results are unlikely to reach clinical standards. Additionally, fully automated grading algorithms could run into acceptance problems, because robustness in a clinical workflow has yet to be proven.
We regard mitosis detection as an intermediate step needed to propose a region of interest that could either be representing the statistics of the complete slide, or, as typically intended, represents the region of highest mitotic activity. For this approach, however, it is valid to not consider the object detection task of mitotic figures, but rather to derive maps where mitotic cells are located. 

\subsection{Mitosis as Segmentation Task}

For the purpose of field of interest proposal, we  consider mitotic figure detection a segmentation task, with mitotic figures being represented by filled circles. This enables the use of concepts like the dice coefficient (intersection/union) for both, evaluation as well as for optimization.

\begin{figure*}[!t]
  \centering
  \includegraphics[width=\linewidth]{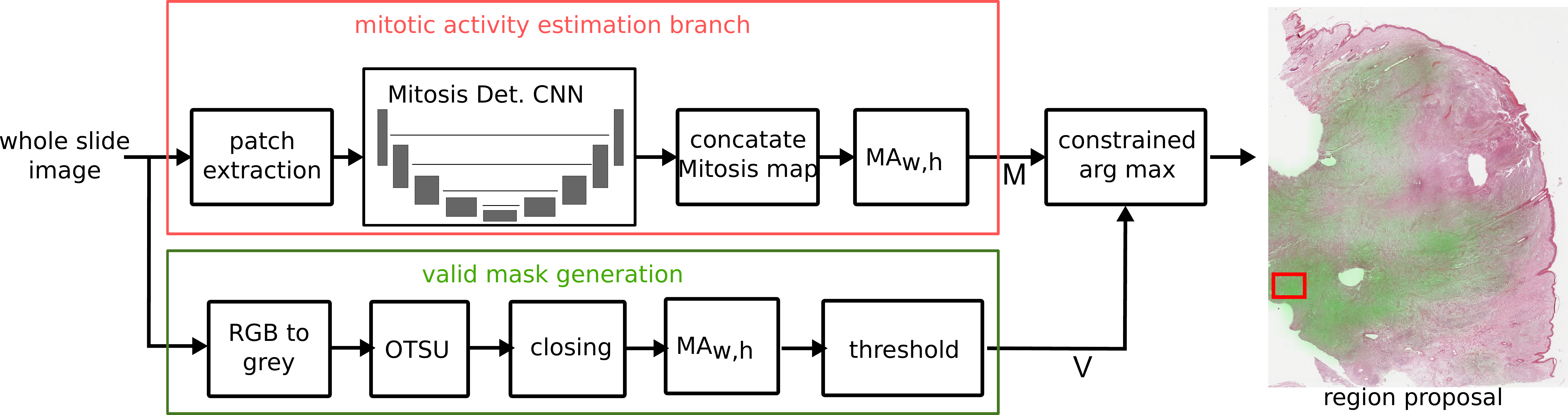}
  \caption{Overview of the proposed approach for mitotic count region proposal. The upper path will derive singular mitotic annotations, followed by a moving average (MA) filter. The lower path derives an activity map of the image to exclude border regions of the image. Region proposal (red rectangle) shows result on slide taken from the test set with ground truth MC depicted as as green overlay.}
  \label{fig:overview}
 \end{figure*}

\subsection{High Power Field Area Proposal}

We employ an approach based on prediction of mitotic activity (upper path in Fig. \ref{fig:overview}) and an estimation of a valid mask (lower path of Fig. \ref{fig:overview}), which will select image areas that are covered to a very large extent by tissue. 
A single HPF at field number 22 is assumed to have an area of $A = 0.237\,mm^2$ \cite{Meuten:2016jh}. In order to find an area with the size of 10 HPF, we thus look for an moving average estimator with the following width $w$ and height $h$ in pixels (aspect ratio of 4/3 is assumed)
 
\begin{equation}
	w = \frac{\sqrt{\frac{10 \cdot 4}{3} A}}{r} \cdot 1000 \frac{[\mu m]}{\textrm{[mm]}}
\hspace{1cm}
	h = \frac{\sqrt{\frac{10 \cdot 3}{4} A}}{r} \cdot 1000 \frac{[\mu m]}{\textrm{[mm]}}
\end{equation}

with $r$ being the resolution of the scanner (in $\mu m/px$). 

\subsubsection{Mitotic Activity Estimation}
For estimation of mitotic activity, the image is divided into overlapping (margin: 64px) patches of size $512\times 512$. The prediction of the network is being concatenated to yield a map of mitotic figure activity $M$.  

\subsubsection{Valid Mask Estimation}

In order to exclude regions of the image that are partly uncovered by specimen, we construct a binary mask of tissue presence from the WSI: The downsampled image is converted to grey-scale, then a binary threshold is performed using Otsu's adaptive method \cite{Otsu:1979hf}. A closing operator is applied to reduce thin interruptions of the tissue map, and a moving average filter of size $w \times h$ is being applied. Next, a thresholding with 0.95 is applied to retain only areas that are covered to at least 95\,\% with tissue, resulting in the valid mask $M$. 
Lastly, both masks are used to find the position of the maximum mitotic activity, constrained to image areas where the valid mask is nonzero.

\subsection{Convolutional Neural Network (CNN) Structure}
We follow the popular U-Net architecture of Ronneberger \textit{et al.} \cite{Ronneberger:2015gk}, which was successful in segmentation tasks in microscopy images, and use a network consisting of five stages, each containing two 2D convolutional layers with batch-normalization followed by a max pooling layer in the downsampling branch. The network then uses an upsampling branch and feeds information of the layers of matching resolution to the upsamling convolutions. Finally, a convolution layer with sigmoid activation function is being used.

The ground truth image map is being generated as filled circles around the actual annotation coordinates of each mitotic figure in the current image patch. This approach follows the original works by Cire\c{s}an \textit{et al.} \cite{Ciresan:2013up}, where the CNN-based detector would receive a positive mitosis indication as ground truth if the closest annotation distance is less than a given radius.  
Following Rahman and Wang \cite{rahman2016optimizing}, we directly use the Intersection over Union (IOU) for binary classification as optimization loss for our task. In heavily imbalanced problems such as our mitosis segmentation task, IOU will yield a balanced measure. We skip the constant term in their formula, and formulate the loss as:

\begin{equation}
L_{IoU} = -\frac{\sum_{v \in V}{X_v+Y_v}}{\sum_{v \in V}{X_v + Y_v - X_v*Y_v}}
\end{equation}

with $V$ being the totality of all pixels and $X_v$ and $Y_v$ being the ground truth and predicted labels at pixel position $v$, respectively. We use the Adam optimizer with an initial learning rate of 0.0005 in TensorFlow.

To split between training and validation set, we perform a vertical split of the 22 WSI, where the lowest 20\% are used for validation. We use random rotation as augmentation. For training, we feed tuples of images to the network, representing three groups of images with different intentions. The first group consists of arbitrary image patches containing at least one mitotic figure. This group is responsible to not have a complete underrepresentation of the positive class in our data set. The second group represents hard negative examples. It consists of images that contain at least one mitosis candidate where both experts disagreed on the cell class or where cells have been classified by both experts as unable to classify. The third group is completely random picks of images on the slide. This group ensures also picking images that do not contain tissue with mitotic figures for images that do not contain a large number of mitoses as well as border or non-tissue regions  on the slide.

\section{Results}

\begin{figure}[!ht]
  %\vspace{-0.2cm}
  \centering
  \subfigure[]{
  \includegraphics[width=0.46\linewidth]{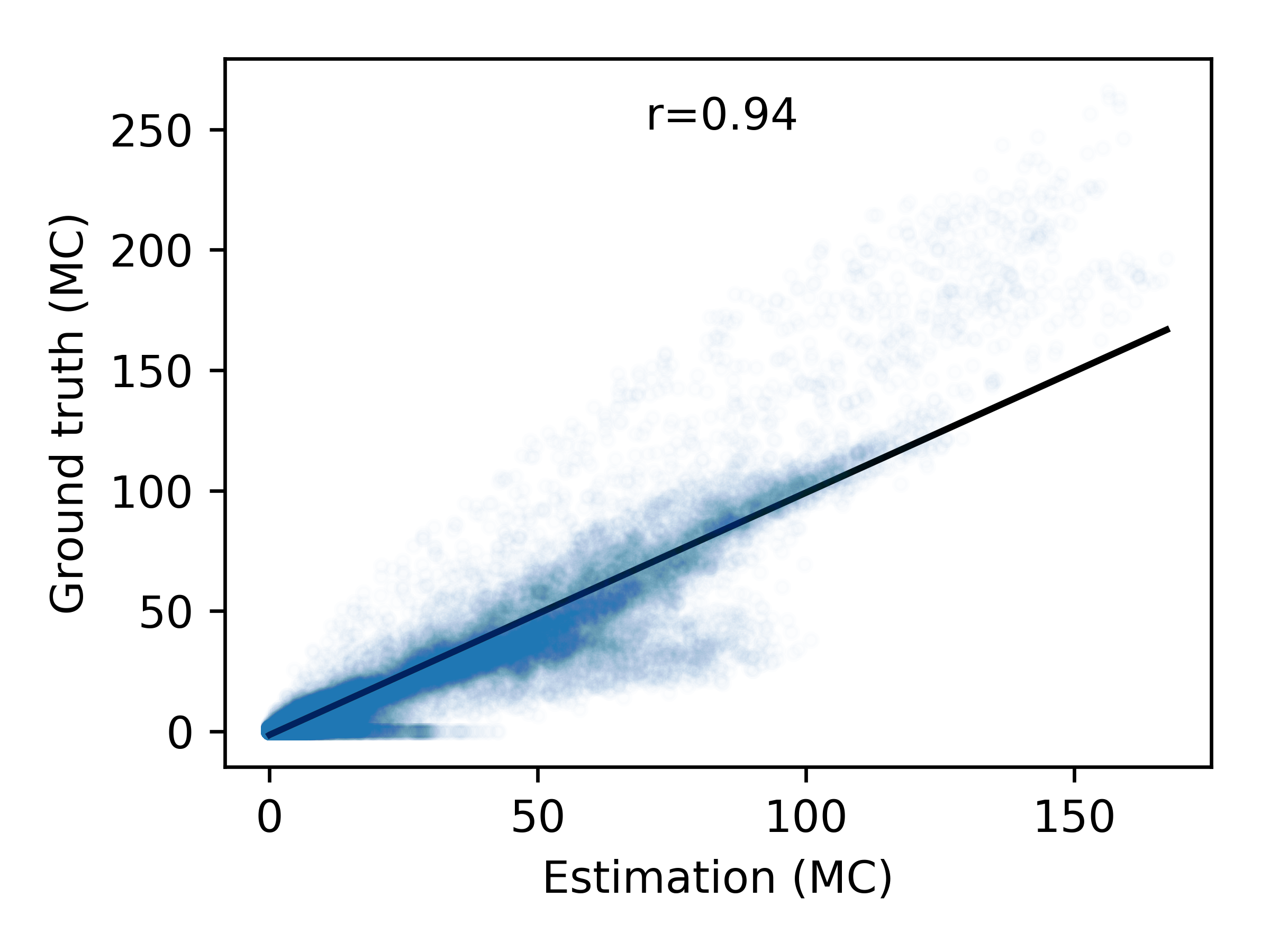}\label{fig:corr}}
  \subfigure[]{
  \includegraphics[width=0.46\linewidth]{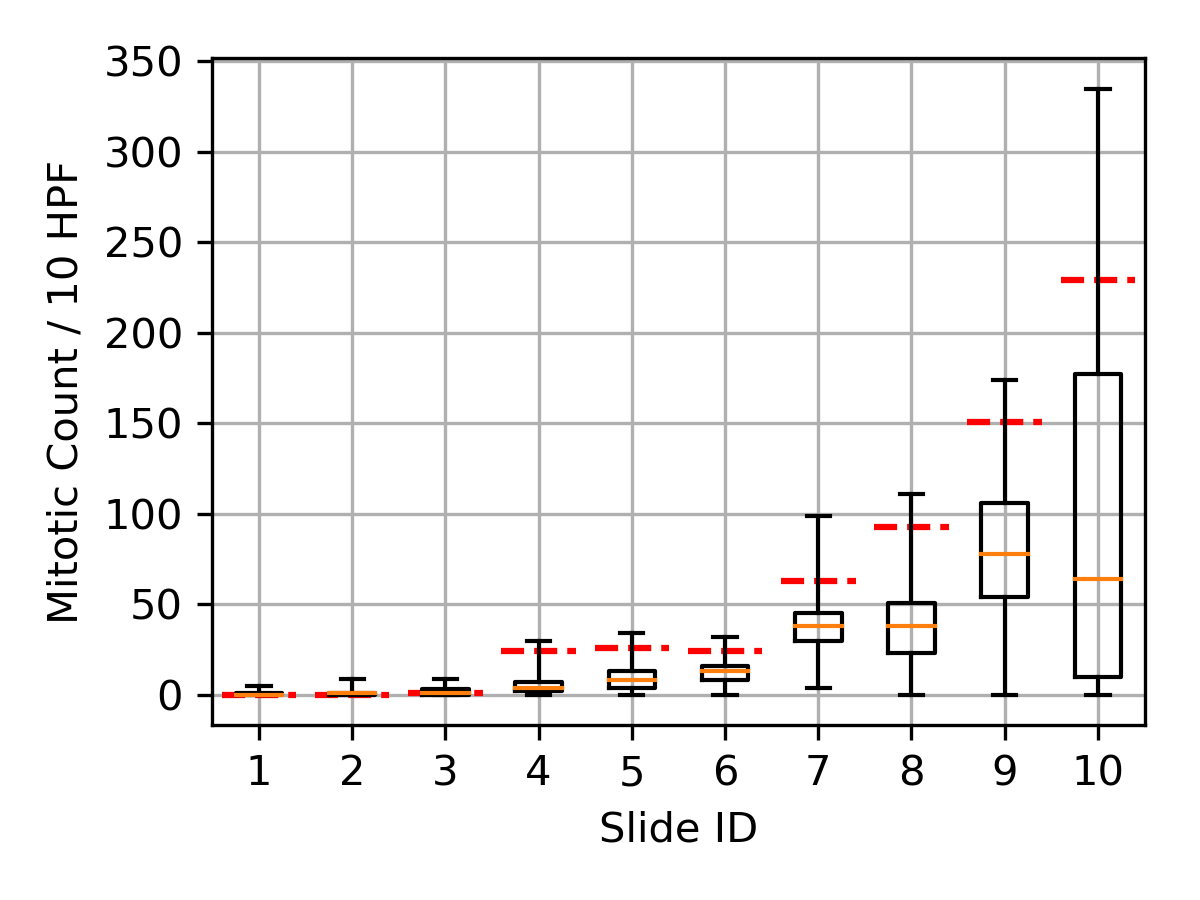}
  \label{fig:distribution}}
  \caption{a) Relation between ground truth mitotic count (MC) prediction and estimated MC prediction on test set (r=0.936) b) MC distribution on test slides (ground truth). Red dashed line marks ground truth MC for proposed position.}
 \end{figure}

On the test set, we achieve a correlation coefficient in mitotic count estimate of $r=0.936$, with partial over-estimation on a small part of the data set (see~Fig.~\ref{fig:corr}). The overall F1-score of the intermediate mitotic figure prediction task of the network is 0.662, the mean IOU is 0.495. 
Our test dataset shows a significant spread of the mitotic count within the specimen of the respective test slides, as indicated by the box-whisker-plot in Fig. \ref{fig:distribution}. Some of the slides~(1~to~3) have low mitotic count, reflecting a low low grade tumor, while others show clearly high-grade tumors~(7-10). The slides with intermediate mitotic counts (4-6) are of special interest, since the ground truth MC ranges closely around the commonly used cut-off value of MC$\geq$7 and thus we would expect a higher variability of the assigned tumor grade if the same slide is assessed by multiple independent experts. The approach presented in this work chose for all relevant slides a position in the forth quartile of the ground truth MC (see red dashed line in Fig. \ref{fig:distribution}, for an example see image on the right of Fig.~\ref{fig:overview}).

\section{Discussion}
The mitotic figure prediction network scored in the same order as other algorithms on other data sets that also do mitosis detection. However, while the general problem of automatically identifying  mitotic figures in WSI with sufficient accuracy for clinical application remains a challenge, the outcomes of these approaches might indeed serve as a surrogate for field of interest proposal and thus as a augmentation to the pathology expert. In future studies, it will have to be proven that clinical application of such augmentation methods will be able to reduce variability in MC determination.

\bibliographystyle{bvm2019}

\end{document}